\documentclass[letterpaper, 10 pt, conference]{ieeeconf}  % Comment this line out if you need a4paper

\IEEEoverridecommandlockouts % This command is only needed if 
\usepackage{graphicx} % Required for inserting images
\usepackage{xcolor}
\usepackage{amsmath}
\usepackage{amsfonts}
\usepackage{graphicx}
\usepackage{subcaption}
\usepackage{cite}
\usepackage{bm}
\usepackage{amssymb}

\title{\LARGE \bf MOCHA: Multi-Objective Co-Design using Hypernetwork Architectures}

\newif\ifanonymous
\ifanonymous
    \author{Anonymous Authors}
\else
    \author{Varun Madabushi, Neil Janwani, and Maegan Tucker% <-this % stops a space
    \thanks{This work is supported by the Georgia Tech Institute for Robotics and Intelligent Machines (IRIM) and NSF (CPS Award \#2440387)}
    \thanks{Authors are with the Dynamic Mobility Lab at Georgia Tech, Atlanta, U.S. \texttt{\{vmadabushi, njanwani, mtucker34\}@gatech.edu}}
    }
\fi

\DeclareMathOperator*{\argmax}{arg\,max}

\begin{document}

\maketitle
\begin{abstract}
    % Current approaches to jointly optimizing robot control and physical design, denoted \textit{robot co-design}, rely on complex, nested optimization which, when combined with long training procedures, makes it challenging to generalize across tasks or reward functions.
    In this work, we present MOCHA, the first, to our knowledge, reinforcement learning based approach to computing a family of Pareto-optimal policies across the design space of a robot using a single network.
    Specifically, MOCHA leverages the hypernetwork architecture to learn a network that produces specialized network parameters optimized for a given objective and parameterized robot design; we term this a multi-objective design hypernetwork (MDH).
    We demonstrate the capabilities of MDHs to represent a complex family of design-dependent strategies on two distinct robot morphologies, each with six design dimensions and across 2-3 objectives.
    Moreover, we propose an approach for efficiently producing a \textit{Design Pareto set} using evolutionary search of the learned policy network, generating  the optimal design-policy combination for each objective prioritization.
    Lastly, we provide an efficient method for computing \textit{generalist robot designs} which achieve the best cumulative performance across the entire set of objectives.
    
    % across a three-dimensional objective space, ultimately discovering the optimal design for each unique scalarization of objectives on two robot morphologies.
    % This hypernetwork policy is trained alongside a design predictor network, which learns to map an objective prioritization to the optimal robot design.
    % Co-training the policy and design predictor improves the policy performance over the pareto set of designs.
\end{abstract}
\section{Introduction}
Robot co-design seeks to holistically consider the coupled effects of mechanical and control parameters on overall performance of robotic systems in an algorithmic fashion.
By jointly optimizing the mechanics and control, co-design algorithms can achieve higher system performance than separately designing each.
Traditional robot co-design treats the relationship between design and control parameters as a bi-level optimization problem, in which the outer loop selects a design and the inner loop selects a controller \cite{ha2018computational,bhatia2021evolution, kim2021mo, yue2025toward}.
However, this architecture poses a challenge when using learning-based control policies, since each candidate design requires its own time-consuming training run.

% This long training time results in a time-consuming trial-and-error process when exploring reward-design tradeoffs. 
% This poses a challenge when the inner loop control design involves training a reinforcement learning (RL) policy.
One simple approach that has been demonstrated for reinforcement learning (RL) based co-design is to train a \textit{universal policy} (UP) conditioned on a description of the robot design \cite{Luo2024Morphologically, feng2022genlocogeneralizedlocomotioncontrollers, bohlinger2025policyrunallendtoend}.
This flattens the bi-level optimization problem into a single loop that searches the design space for high-quality designs, guided by the policy rollouts or a universal value function.
However, this results in policy behavior being tied to a specific pre-defined reward function.
% Notoriously, the performance of an RL policy is sensitive to the structure and relative weights of the various sub-objectives within the reward function.
Once trained, the agent can only express behavior consistent with the reward function it was trained on, posing a challenge when the task prioritization is not exactly known beforehand.

\begin{figure}
    \centering
    \includegraphics[width=\linewidth]{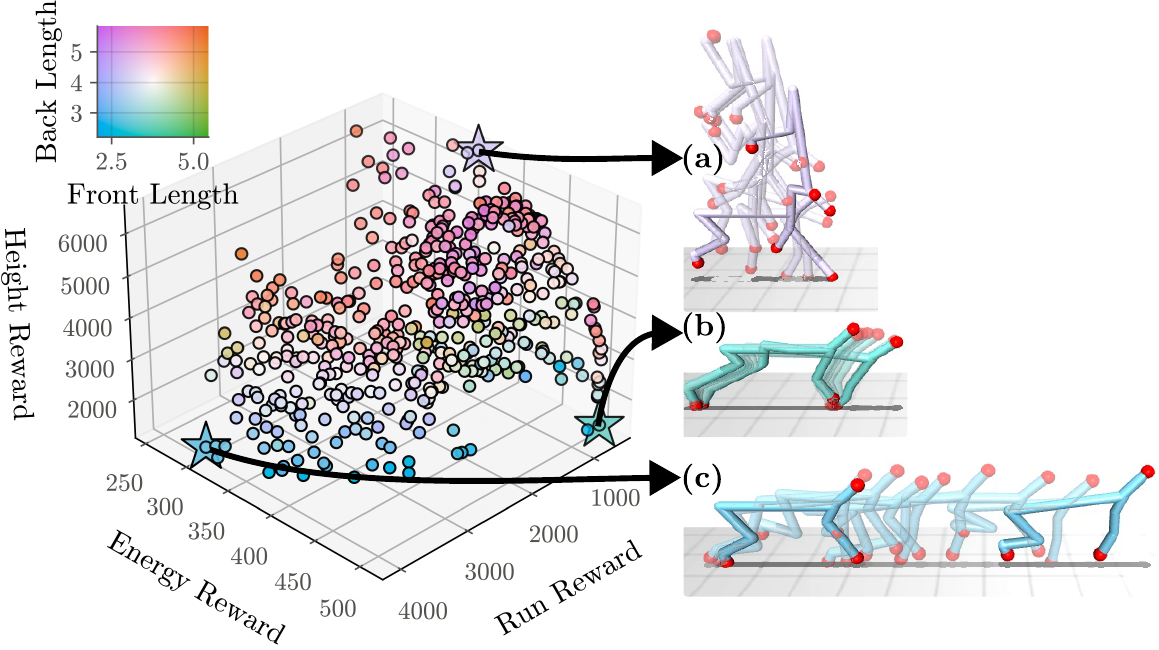}
    \caption{By learning a control policy over multiple objectives and designs, our algorithm is capable of finding the optimal design for a given objective prioritization. For example, (a) is optimized for jumping, (b) is optimized for energy consumption, and (c) is optimized for running speed.}
    \label{fig:hero}
\end{figure}

In contrast, Multi-Objective RL (MORL) measures performance across multiple objectives with a vector-valued reward function.
The goal of MORL is to learn a Pareto set of policies that optimally trade off across objectives.
Explicitly, policies are only included in the Pareto set if no other policy can outperform it on an objective without sacrificing performance in another.
By considering a Pareto set of policies instead of a single policy, one can explore the space of possible behaviors and train a single agent to complete multiple tasks without retraining. 
However, MORL has only been applied to robots with a fixed morphology and has not yet been extended to design-conditioned policies. 

Our approach, titled MOCHA, combines the benefits of MORL and design-conditioned policies to train a \textit{Multi-Objective Universal Policy} (MO-UP), whose value function and control policy optimally trade-off across multiple objectives.
We argue that this combination unlocks the true potential of co-design by allowing a designer to explore the relationship between design and performance across competing objectives, and appropriately manage trade-offs. 

We construct the MO-UP as a pair of value and policy hypernetworks, motivated by their success in expressing diverse families of policies in Multi-Objective Learning and Meta-Learning \cite{janwani2026moplaygroundmassivelyparallelizedmultiobjective, beck2022hypernetworksmeta}. 
Rather than representing the entire family of behaviors with a single fixed set of network parameters, a hypernetwork generates the parameters of a policy or value network conditioned on a context vector.
This provides a parameter-efficient mechanism for representing a continuous family of specialized controllers across robot designs and objective trade-offs.
We call our specific class of hypernetwork a \textit{Multi-Objective Design Hypernetwork} (MDH).
MOCHA also introduces an algorithm to search the input space of the MDH, producing a \textit{Design Pareto frontier}, that is, a Pareto frontier of design-policy pairs, demonstrating its value in design space exploration.

The individual contributions of this work are as follows:
\begin{enumerate}
    \item A demonstration of the hypernetwork architecture's capability to represent a family of Multi-Objective policies across the design space of a robot.
    % \item A training procedure that encourages diversity and performance among strategies taken by the hypernetwork policy.
    \item An algorithm that uses a hypernetwork as the Multi-Objective Universal Policy to compute a Pareto set of design-policy pairs.
    \item An algorithm for efficiently finding a \textit{generalist} robot design, which achieves the best cumulative performance on all objectives.
    \item An open-source, GPU-compatible implementation of the algorithms and benchmarks for Multi-Objective Co-design.
\end{enumerate}

\section{Related Work}
Our work applies concepts from MORL and Multi-Domain RL to generate a policy that can be re-used for any design within a parameterized design space. 
This policy is embedded in a co-design algorithm to generate a Pareto set of designs.
In this section, we will discuss existing methods to solve related problems.

\subsection{Multi-Objective Reinforcement Learning}
% \textcolor{red}{We will discuss the MORL algorithms such as AMOR, HyperMORL, MORLAX}.
Unlike traditional RL, which maximizes a single scalar reward, MORL instead considers a vector-valued reward function consisting of a number of reward components.
In most cases, the reward components cannot all be simultaneously maximized, as there are trade-offs between the objectives.
Thus, MORL learns a Pareto set of policies, which optimally trade off across the objectives.
% \textcolor{red}{Should I include the formula/definition of Pareto optimality as in MO-Playground?}.

A number of algorithms exist to learn a continuous representation of the Pareto set.
AMOR \cite{Alegre2025AMOR} and PD-MORL \cite{basaklar2023pdmorl} both achieve this by learning a single policy (represented as a neural network) conditioned on the trade-off.
However, both suffer from long training times, potentially due to the challenge of using a single neural network to represent both a policy distribution over the complex set of trade-offs and a robust control strategy.

Researchers have turned to hypernetworks, neural networks that output other networks conditioned on a context vector, as an alternative to represent the continuous Pareto set of policies.
HYPER-MORL \cite{shu2024learningparetosetmultiobjective}, PSL-MORL \cite{liu2025paretosetlearningmultiobjective}, and MORLAX \cite{janwani2026moplaygroundmassivelyparallelizedmultiobjective} demonstrate that hypernetworks are capable of representing these policy families in a parameter-efficient manner.
MORLAX connects hypernetwork-based MORL with GPU acceleration in JAX \cite{jax2018github} to train Multi-Objective Policies using a Multi-Objective formulation of Proximal Policy Optimization (PPO).
In this work, we build upon the MORLAX algorithm to train a multi-objective policy hypernetwork across the design space of a robot.

\subsection{Multi-Domain Reinforcement Learning}
% \textcolor{red}{Discuss works like the Bohlinger UP, the Dudek paper.}
The goal of Multi-Domain RL is to train a policy on a set of domains and achieve zero- or few-shot transfer to domains not seen in training.
In the context of robot co-design, this enables a single reusable policy to be used to evaluate different robot designs.

Luo et al. and Feng et al. train controllers to generalize to a class of quadruped robots through randomization of the morphological parameters (mass and length) \cite{Luo2024Morphologically}, \cite{feng2022genlocogeneralizedlocomotioncontrollers}. This aligns with the broader paradigm of Hardware-Conditioned Policies (HCPs), where a single neural network takes explicit physical parameters (e.g., link masses, lengths) as inputs to adapt its control strategy across diverse robot configurations \cite{yu2017preparing, chen2018hardware}.
In these settings, the robots are limited to a fixed kinematic structure, keeping the observation and action space constant across designs.
To resole this, Bohlinger et al. leveraged an encoder-decoder architecture with a shared backbone to accommodate robots with differing observation and action spaces \cite{bohlinger2025policyrunallendtoend}.

Hypernetworks have also demonstrated effectiveness in multi-domain generalization, particularly in the related problem setting of meta-learning.
Rezaei-Shoshtari et al. used supervised learning to train a hypernetwork on a dataset of trajectories obtained by sampling pre-trained RL policies, enabling zero-shot transfer to unseen tasks and reward weights \cite{rezaeishoshtari2023hypernetworkszeroshottransferreinforcement}.
Additionally, Beck et al. trained a recurrent hypernetwork in an end-to-end fashion to represent a family of solutions to a Meta-RL problem \cite{beck2023recurrenthypernetworks}.
The authors demonstrate that hypernetworks can prevent interference between distinct tasks while sharing parameters as necessary, resulting in diversity and parameter efficiency; properties which are leveraged in our work.

\begin{figure*}[ht!]
    \centering
    \includegraphics[width=\linewidth]{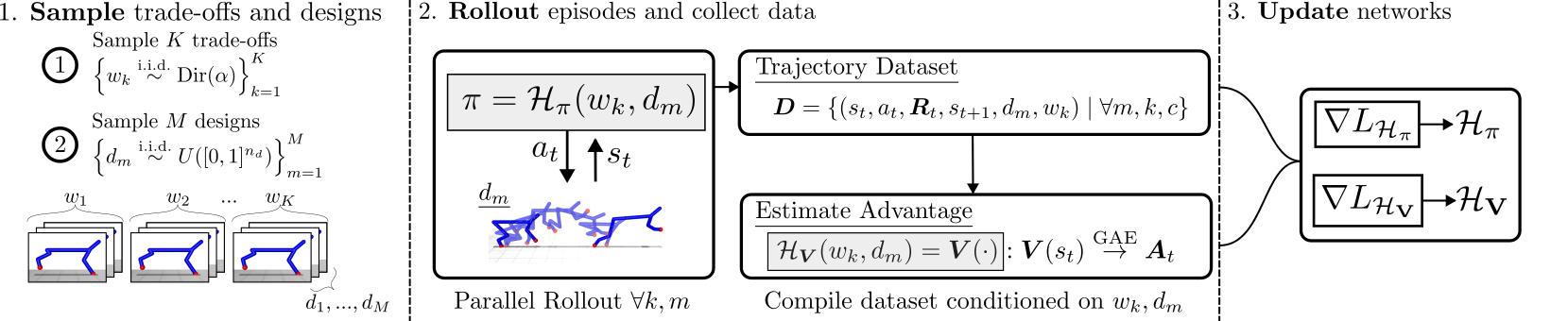}
    \caption{The MO-UP Hypernetwork is trained by first sampling $M$ designs and $K$ trade-offs.
    The rewards obtained from rolling out these designs and trade-offs are used to update the value and policy networks using Multi-Objective PPO.}
    \label{fig:method}
    % \vspace{-0.5cm}
\end{figure*}

\subsection{Co-Design}
Jointly selecting mechanical and control parameters of a robot is a well-studied problem, beginning with bilevel optimization-based approaches \cite{ha2018computational, ghansah2023humanoidrobotcodesigncoupling}.
Recent work formulates the co-design problem directly as an RL problem.
These include leveraging evolutionary algorithms \cite{gupta2021embodied, bhatia2021evolution}, or a co-trained designer policy \cite{yuan2022transform, schaff2022nlimb} to optimize morphology while using RL to train the control policy.
Other methods utilize graph neural network-based heuristics \cite{zhao2020robogrammar}, or an RL policy's value function \cite{bohlinger2026shapebodyvaluegradients, luck2019coadaptation}, as a proxy for design quality in a downstream design search algorithm.
These approaches all use networks which take the design as input and are limited to single-objective learning tasks.

One of the few examples of multi-objective co-design is Multi-Objective Graph Heuristic search, which integrates multi-objective learning with co-design to analyze the Pareto set of robot designs \cite{Xu2021MOGHS}.
This approach learns a heuristic function inspired by critic networks in RL to guide the design search, but expresses the control policy using model-predictive control.
Similarly, Kim et al. proposed Multi-Objective Bilevel Bayesian Optimization (MO-BBO) to capture the Pareto set of robot morphologies and behaviors, though their approach relies on Bayesian optimization and has not been demonstrated for high-dimensional control parameterizations.
In our work, we investigate hypernetworks as an alternative to design-conditioned policy networks in order to enable multi-objective robot co-design with RL-based control.

\section{Method}
\subsection{Problem Setup}
% \textcolor{red}{The notation here can be expanded upon/improved.
% Especially an explanation of the scalarization used in multi-objective learning}
Consider a normalized design space $\mathcal D=[0, 1]^{n_d}$ representing the $n_d$-dimensional space of valid robot morphology parameters.
We generalize the Multi-Objective Markov Decision Process (MOMDP) framework from MORL to be a design-aware MOMDP (DA-MOMDP) $(\mathcal{S}, \mathcal{A}, \mathcal{T}, \mathcal{D}, \bm{R}, \gamma)$, where $\mathcal{S}$ is a set of states, $\mathcal{A}$ is a set of actions, $\mathcal{T}(s' \mid s, a, d)$ is the probability of transitioning from state $s \in \mathcal{S}$ to state $s' \in \mathcal{S}$ given action $a \in \mathcal{A}$ and design $d \in \mathcal{D}$,
and $\gamma \in [0,1)$ is a discount factor.

The vector-valued reward function $\boldsymbol{R}: S \times A \rightarrow \mathbb{R}^{n_r}$ is composed of $n_r$ reward components, each describing a different objective.
The trade-off vector $w$ represents a relative prioritization of reward components and is drawn from $\Delta^{n_r-1}$, the $(n_r-1)$-dimensional simplex.

\subsection{Hypernetwork Architecture}
We represent the family of policies and value functions across the design and trade-off space using two hypernetworks: the Actor hypernetwork and Critic hypernetwork.
Respectively, 
\begin{align}
    \mathcal H_\pi: \Delta^{n_r - 1} \times \mathcal D \rightarrow \Theta_\pi, \quad  
    \mathcal H_V: \Delta^{n_r - 1} \times \mathcal D \rightarrow \Theta_V, 
\end{align}
which learn to predict the parameters of the policy and value neural networks with parameter spaces $\Theta_\pi$ and $\Theta_V$, respectively.
These functions are represented in the affine form
\begin{gather}
    \mathcal H_{\pi, V}(w, d) = W_{\pi, V}f_{\pi, V}(w, d) + b_{\pi, V}
\end{gather} 
where $f_{(\cdot)}$ is a nonlinear function (represented as a neural network) termed the \textit{task encoder} that maps a context vector comprised of $w$ and $d$ into a higher-dimensional task embedding.
The task embedding is then transformed into the parameter space of the base networks through the matrix $W_{(\cdot)}$ and column vector $b_{(\cdot)}$.
When successfully trained, this architecture learns a lower-dimensional manifold on which the set of relevant policies live, and indexes policies from this manifold using the corresponding context.

\subsection{Multi-Objective Reinforcement Learning}
The goal of MORL is to learn a policy that maximizes $w^T\boldsymbol{J}_\pi$, where
\begin{gather}
\boldsymbol{J}_\pi=\mathbb{E}_{s_0 \sim \mathcal{S}_0} [\boldsymbol{V}_\pi(s_0) ] \in \mathbb{R}^{n_r}
\end{gather}
$\boldsymbol{J}_\pi$ represents the expected return of policy $\pi$ over initial state distribution $\mathcal{S}_0$,
and $\boldsymbol{V}_\pi(s_0)$ is the vector-valued value function evaluated at the initial state $s_0 \in \mathcal{S}_0$, defined as 
% \begin{gather}
%     \boldsymbol{V}_\pi(s) = \mathbb{E}\left[ \sum_{t=0}^\infty \gamma^t\boldsymbol{R}(z_t, \pi(z_t))  \mid z_t \sim \mathcal{T}(z_{t-1}, \pi(z_{t-1}), d), z_0 = s \right] \in \mathbb{R}^{n_r}. 
% \end{gather}
\begin{gather}
    \begin{split}
        \boldsymbol{V}_\pi(s) = \mathbb{E}\biggl[ &\sum_{t=0}^\infty \gamma^t\boldsymbol{R}(z_t, \pi(z_t)) \\
        &\qquad \mid z_t \sim \mathcal{T}(z_{t-1}, \pi(z_{t-1}), d), z_0 = s \biggr] \in \mathbb{R}^{n_r}.
    \end{split}
\end{gather}

The formulation of this learning problem mirrors that of single-objective RL, except that the reward and value functions are vector-valued.
The policy is learned through a multi-objective extension of the Proximal Policy Optimization algorithm \cite{Xu2020pgmorl}.

Since it may not be possible for a single policy to maximize all components of $\boldsymbol{J}_\pi$, the goal is to instead find a set of Pareto-optimal policies which optimally trade-off the objectives.
Pareto optimality is determined by the notion of \textit{Pareto dominance}: a policy $\pi$ Pareto dominates a policy $\pi'$ if $\boldsymbol{J}_\pi \geq \boldsymbol{J}_{\pi'}$ across every component, and $\boldsymbol{J}_\pi \neq \boldsymbol{J}_{\pi'}$ for at least one component.
A policy is \textit{Pareto optimal} if it is not dominated by any other policies.
The set of all such optimal policies forms the \textit{Pareto set}, and their corresponding rewards make up the \textit{Pareto frontier}.

We measure the quality of a Pareto frontier by computing its \textit{hypervolume} \cite{zitzler1998multiobjective}, defined as the volume enclosed by the Pareto frontier with respect to a given reference point (taken to be $\mathbf{0}$ in our implementation), and denote it as $\Gamma: \Pi \rightarrow \mathbb{R}$, where $\Pi \subset \Theta_\pi$ is a set of policies.
Note that because we choose the reference point to be the origin, all rewards must be positive.
We also measure the \textit{spacing} of the Pareto frontier, which is the standard deviation of the distance between each point on the Pareto frontier and its nearest neighbor (in the reward space).
A high quality Pareto frontier has high hypervolume and low spacing.

\subsection{Multi-Objective Design-Conditioned Hypernetwork}
We propose a Multi-Objective Design-Conditioned Hypernetwork (MDH) architecture to represent a MO-UP, trained through the following procedure, and illustrated in Figure \ref{fig:method}.
This approach builds upon MORL by training a policy to maximize $w^T\boldsymbol{R}$ over every design in $\mathcal D$.

\subsubsection{Initialization}
The algorithm is initialized with $M$, the number of sampled designs, $K$, the number of sampled trade-offs, and $C$, the number of trials for each trade-off-design pair.
The MDH weight matrices $W_{(\cdot)}$ are initialized to all zeros, and the weight vectors $b_{(\cdot)}$ are initialized randomly using Kaiming initialization \cite{he2015delvingdeeprectifierssurpassing}.
The weights of the task encoders are also initialized with Kaiming initialization.
Empirically, we find that this initialization helps to promote diversity in the hypernetwork policies, corroborating \cite{beck2022hypernetworksmeta}.

\subsubsection{Sampling and Rollout}
Each training epoch begins with drawing a set of $K$ trade-offs $\{w_k\}_{k=1}^K$.
The first $K-n_r$ are sampled from a Dirichlet distribution with concentration parameters $\alpha_i=1, i=1,...,n_r$, and the remaining are taken to be the corners of $\Delta^{n_r-1}$, (e.g. for $n_r=3$, $[1, 0, 0], [0, 1, 0], [0, 0, 1]$).
This guarantees that the extreme tradeoffs are explored.
Then, $M$ designs $\{d_m\}_{m=1}^M$ are sampled uniformly from $\mathcal{D}$ using a Sobol sequence \cite{SOBOL196786}.
This grid of design-trade-off combinations is used to instantiate $M\cdot K\cdot C$ parallel environments.

Each environment is simulated according to its dynamics (parameterized by $d_m$) and the policy $\pi_{w, d} = \mathcal{H}_\pi(w_k, d_m)$.
These rollouts are collected into a dataset $\bm{D}$ of design-conditioned state-action transitions, with each entry of the dataset defined as 
\begin{gather}
    \boldsymbol{D}^{(m, k, c)}= \left(s_t^{(m, k, c)}, a_t^{(m, k, c)}, \boldsymbol{R}_t^{(m, k, c)}, s_{t+1}^{(m, k, c)}, w_k, d_m\right),
\end{gather}
and $m,k,$ and $c$, denoting the design, trade-off, and trial indices, respectively.

\subsubsection{Gradient Update}
The collected dataset $\bm{D}$ is used to update the MDH weights.
First, we estimate a vector-valued advantage for each state-action trajectory $\boldsymbol{A}_t \in \mathbb{R}^{n_r}$ by applying Generalized Advantage Estimation independently to each objective \cite{schulman2018gae}.
Traditionally, these advantages are shuffled, grouped into batches, and normalized by the batch mean and standard deviation before being used to compute the policy gradient.

We observe that naively normalizing the advantages without regard for the design results in mode collapse, where the policy learns to take the same or similar actions irrespective of the design.
To avoid this, we group the set of advantages according to their associated design, and normalize each advantage relative to only the others in its group.
The group-wise normalized advantages are used to update the hypernetworks by applying stochastic gradient descent to minimize the following losses (dataset indices $(m,k,c)$ dropped for readability):
\begin{align}
L_{\mathcal{H}_\pi} &=
-\mathbb{E}_{(s_t,a_t,w, d) \sim \boldsymbol{D}}
\left[
r_t(\theta)\, w\cdot \boldsymbol{A}_t
\right],
\\
L_{\mathcal{H}_{V}}
&=
\mathbb{E}_{(s_t,w, d) \sim \boldsymbol{D}}
\left[
\left\|
\bm{V}_{\phi}(s_t) - \hat{\bm{V}}_t
\right\|_2^2
\right]
\end{align}
\begin{gather}
\text{where }
r_t(\theta)
=
\frac{\pi_{\theta}(a_t \mid s_t)}
     {\pi_{\theta,\mathrm{old}}(a_t \mid s_t)}, \\
     \theta = \mathcal{H}_\pi(w, d), ~\phi = \mathcal{H}_{V}(w, d).
\end{gather}
Here, $r_t(\theta)$ denotes the probability ratio of a particular action taken by the policy with parameters $\theta$ obtained from the hypernetwork with the associated context $(w, d)$, $\bm{V}_{\phi}$ denotes the estimate produced by the value network with parameters $\phi$ also obtained from the hypernetwork, and $\hat{\bm{V}}$ denotes the value computed from the rollouts in the dataset.
In practice, the clipped surrogate PPO loss is used to stabilize training, as in standard PPO implementations \cite{freeman2021brax}.

\begin{figure}
    \centering
    \includegraphics[width=0.7\linewidth]{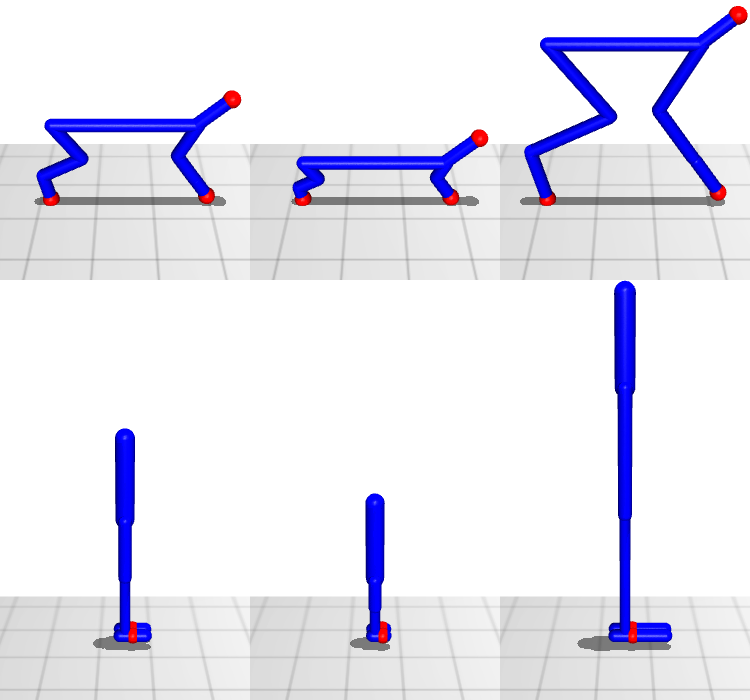}
    \caption{Some examples of procedurally generated  Cheetah and Walker robots.}
    \label{fig:design_examples}
\end{figure}

\subsection{Computing the Design Pareto Frontier}
\label{sec:pareto}
We wish to compute the \textit{Design-Pareto Set}, $\mathcal{P}$, which associates each trade-off
$w^\star \in \Delta^{n_r-1}$ with the design $d^\star$ and conditioning trade-off
$\tilde{w}$ whose induced policy performs best under $w^\star$:
%
% \begin{equation}
%     \mathcal{P} = \Bigl\{(\hat{w}, \tilde{w}, d) \mid (\tilde{w}, d) \in \argmax_{\tilde{w},d} \hat{w}^T \cdot R(w, d)\Bigr\}
% \end{equation}

% \begin{equation}
% \begin{aligned}
%     \mathcal{P} &= \bigl\{ (\hat{w},\, \tilde{w}^\star,\, d^\star) : \hat{w} \in \Delta^{n_r-1} \bigr\}, \\
%     (\tilde{w}^\star, d^\star) &\in \argmax_{\substack{w \in \Delta^{n_r-1} \\ d \in \mathcal{D}}} \; \hat{w}^\top R\bigl(\mathcal{H}(w, d)\bigr).
% \end{aligned}
% \end{equation}
% %
\begin{equation}
\begin{aligned}
    \mathcal{P} = \Bigl\{ (w^\star,\, \tilde{w},\, d^\star) \;\Bigm|\;
        &w^\star \in \Delta^{n_r-1} \;\textrm{ and } \\
        &(\tilde{w}, d^\star) = \argmax_{w, d} \;
            (w^\star)^\top R\bigl(\mathcal{H}(w, d)\bigr) \Bigr\}.
\end{aligned}
\end{equation}

Here $R: \theta_\pi \rightarrow \mathbb{R}^{n_r}$ denotes the vector of expected cumulative rewards obtained
by policy $\pi$. 
We deliberately search over the conditioning weight $\tilde{w}$ independently of the evaluation weight $w^\star$. 
For a given $w^\star$, the policy $\mathcal{H}(\tilde{w}, d)$ obtained from some $\tilde{w} \neq w^\star$ may attain a higher scalarized return $(w^\star)^\top R$ than the policy produced by $w^\star$, $\mathcal{H}(w^\star, d)$. 
A hypernetwork trained to optimality under the MO-PPO objective would theoretically satisfy $\tilde{w} = w^\star$, since conditioning on $w^\star$ would, by definition, yield the policy maximizing $(w^\star)^\top R$. 
In practice this holds approximately, and the conditioning $\tilde{w}$ that produces the best performing policy is often nearly identical to $w^\star$.

While $\mathcal{P}$ is by construction a continuous set over $w^\star$, analytically computing it is infeasible. Prior approaches approximate $\mathcal{P}$ via random sampling \cite{Alegre2025AMOR, janwani2026moplaygroundmassivelyparallelizedmultiobjective}, however this quickly becomes computationally infeasible in moderate to high dimensional design and tradeoff spaces.
Instead, we employ evolutionary search to find a dense collection of 3-tuples $(w^\star, \tilde{w}, d^\star)$ that approximate $\mathcal{P}$.

Specifically, we utilize Non-dominated Sorting Genetic Algorithm (NSGA) III \cite{nsga3}. 
% Briefly, the algorithm works by first defining a set of \textit{reference directions} that span the objective space.
% These reference directions are then used to compute genetic fitness by ensuring that samples remain close to these references and thus remain spread out over the entire front. 
NSGA-III is a computationally-efficient, evolutionary multi-objective optimization program that finds high-quality Pareto fronts by optimizing for both hypervolume and coverage of the objective space.
During optimization, NSGA-III samples points in batches, making it an ideal candidate algorithm for our approach, which already leverages massively parallelized simulation. 
Unlike classic multi-objective optimization where the sample space is restricted to a hyperrectangle, our algorithm searches over the design and trade-off space, where the trade-off is restricted to the simplex.
Thus, we implement a simplex repair method, which projects sampled points back onto the simplex after genetic mixing.
When combined with parallelized simulation, we achieve denser, higher quality Design Pareto sets than by random sampling.

In addition to using the Design Pareto front to identify the best design $d^\star$ and control strategy $\mathcal{H}(\tilde{w}, d^\star)$ for a particular trade off $w^\star$, we also utilize its hypervolume as an indicator of algorithm performance and confirmation of convergence (i.e. when $\Gamma(\mathcal{H}_\pi)$ plateaus).

\begin{figure}
    \centering
    \includegraphics[width=\linewidth]{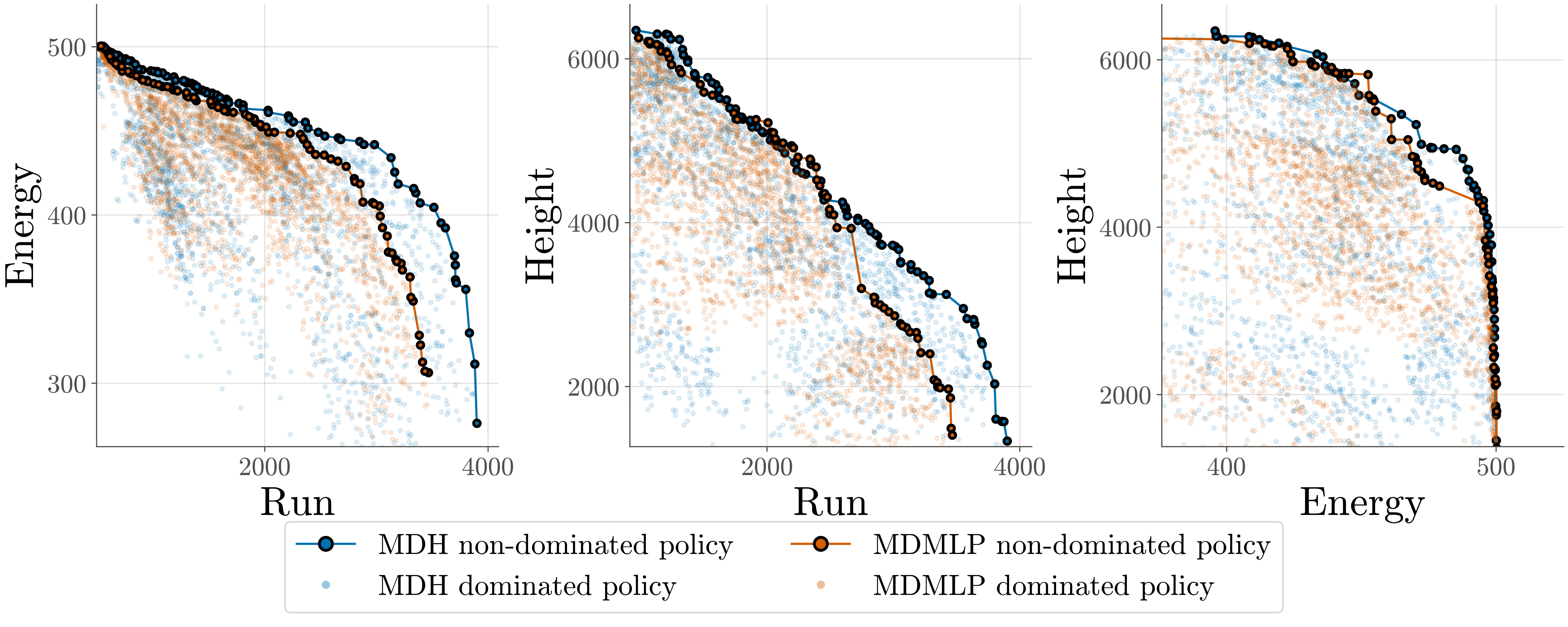}
    \caption{The Design Pareto frontier of the Cheetah sliced along the Run-Energy, Run-Height, and Energy-Height axes.}
    \label{fig:paretocheetah}
\end{figure}

\begin{table}[t!]
\centering
\begin{tabular}{c|c|c|c|c}
    Network & Env. & Hypervolume & Spacing & Train Time \\ \hline
    MDH & Cheetah & \textbf{6.838e9} & \textbf{5.195e-2} & \textbf{2h 6m} \\
    MDMLP & Cheetah & 5.353e9 & 0.1224 & 6h 20m\\
    MDH & Walker  & \textbf{2.715e6} & \textbf{5.656e-2} & \textbf{1h 48m}\\
    MDMLP & Walker  & 2.576e6 & 9.566e-2 & 5h 55m\\
\end{tabular}
\caption{Performance of the hypernetwork algorithm on Walker and Cheetah Environments compared to the MLP baseline}
\label{table:results}
\end{table}

\subsection{Finding a Generalist Design}
\label{sec:generalist}
After applying the method in Sec. \ref{sec:pareto}, points on the Design Pareto frontier correspond to design-policy combinations that maximize a particular objective prioritization.
However, there are situations where a generalist robot design, capable of performing well on all objectives, may be preferred.
For example, if only one robot can be fabricated before being sent on an unknown mission, that design should be capable of handling a large volume of potential missions.

Thus, we define a generalist design as one whose associated multi-objective policy has a high hypervolume.
In other words, it achieves high reward for a large number of possible trade-offs.
We then formulate an optimization problem to find the single design $d^g$ whose associated multi-objective policy has the highest hypervolume.
Formally,
\begin{gather}
d^g =\underset{d\in \mathcal D}{\textrm{argmax}} ~ \Gamma(\mathcal{H}_\pi(w, d)).
\label{eqn:generalist}
\end{gather}
Computing the Design Pareto set's hypervolume is computationally expensive and non-differentiable due to the need to forward-simulate the policy across many sampled tradeoffs and designs.
Therefore, we solve the above optimization problem using Trust Region Bayesian Optimization (TuRBO), a sample-efficient, zeroth-order optimizer \cite{eriksson2020turbo}.

\section{Results}
To analyze the performance of MOCHA, we ask three guiding questions.
First, how good is the MDH at representing optimal policies for any given design-trade-off pair?
Second, does the hypernetwork architecture offer improvements over a multi-objective, design-conditioned multi-layer perceptron (MLP) policy?
Third, how can a practitioner interpret the resulting Pareto set to understand the design space?

\begin{figure}
    \centering
    \includegraphics[width=\linewidth]{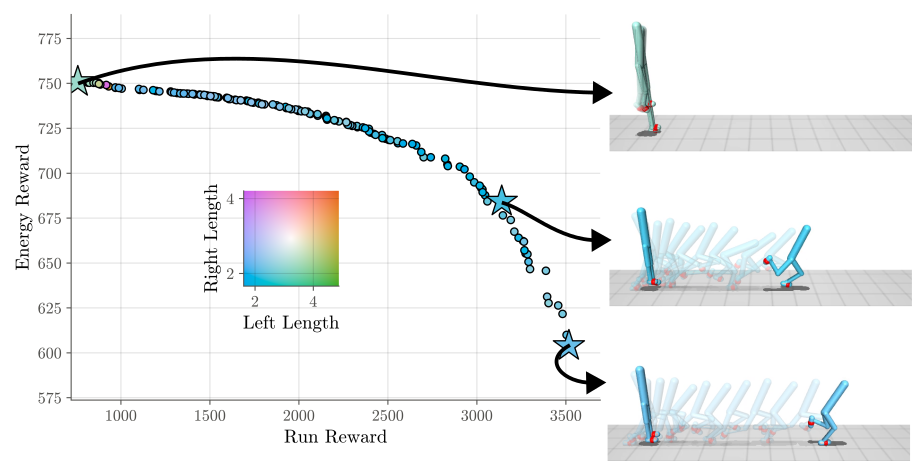}
    \caption{Design Pareto frontier for the 2-objective Walker, and some rollouts of Pareto-optimal designs.}
    \label{fig:paretowalker}
\end{figure}
\subsection{Implementation Details}
We evaluate our approach on a class of robots based on environments from the well-known \texttt{dm\_control} suite \cite{tunyasuvunakool2020}.
In our open-source codebase\footnote{https://anonymous.4open.science/r/codesign-E7CD/}, we implement procedurally generated versions of the Cheetah and Walker environments, some examples of which are shown in Figure \ref{fig:design_examples}.

For our Cheetah environment, we define a 6-dimensional design space corresponding to each of the segments making up the front and back legs.
The 6D normalized design space is affinely mapped onto the range $[0.5, 2]^6$ corresponding to a scaling of of each link relative to the link length of the original robot.
In this environment, the reward consists of three components: (1) \textit{Run}, a reward on the velocity in the +x direction, (2) \textit{Height}, a reward on the height of the robot's hip, and (3) \textit{Energy}, a reward for minimizing the total actuator force.
Similarly, the Walker environment is comprised of a 6D design space corresponding to the lengths of the left and right leg's thigh, shin, and foot segments, formulated using the same scaling procedure as the Cheetah.
The Walker receives Run and Energy rewards, formulated similarly to the Cheetah's.

We implement these environments using the MO-Playground library \cite{janwani2026moplaygroundmassivelyparallelizedmultiobjective}, which leverages MuJoCo JAX \cite{mujoco_playground_2025} for massive parallel computation.
The Pareto frontier metrics (spacing and hypervolume) and NSGA-III algorithm are computed using the \texttt{pymoo} library \cite{pymoo}, and TuRBO is implemented using the \texttt{botorch} library \cite{balandat2020botorch}.

We train an MDH on each of these environments.
The task encoder is comprised of 2 hidden layers of size $[1024, 1024]$ and maps the context (design and tradeoff) to a 128-dimensional task embedding.
The policy and value hypernetworks each have 3 hidden layers of size $[256, 256, 128]$.
Training was conducted on an NVIDIA RTX Pro 6000 Blackwell GPU.
Further details can be found in the linked code repository.

\begin{figure}
    \centering
    \includegraphics[width=\linewidth]{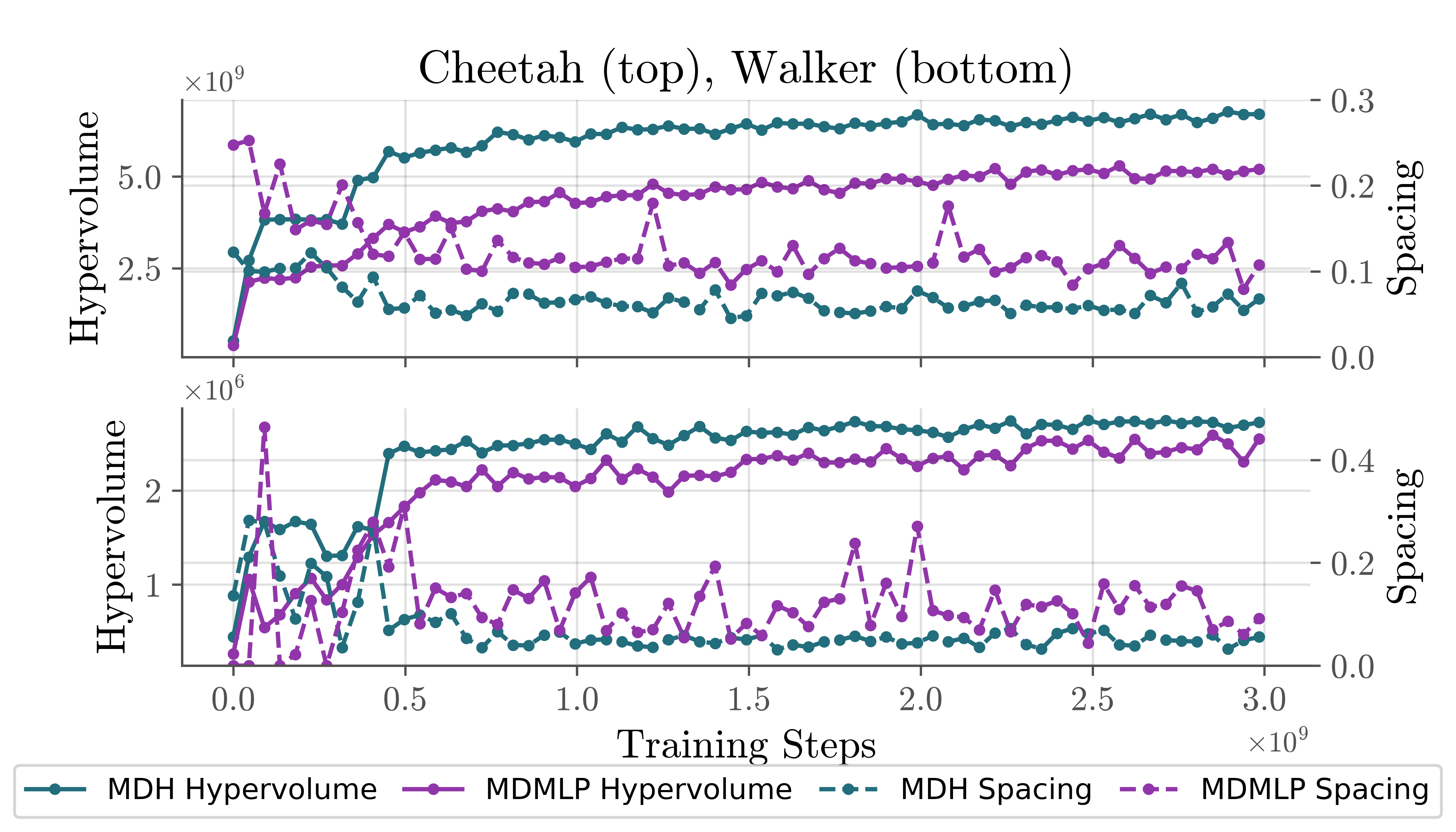}
    \caption{Comparison of the Design Pareto Frontier metrics attained by a Hypernetwork and MLP policy for the Cheetah (top) and Walker (bottom) environments.
    The Hypernetwork achieves a higher hypervolume and lower spacing than the MLP in both environments.}
    \label{fig:training}
\end{figure}

\subsection{Analysis of the Design Pareto Frontier}

The 3D Design Pareto frontier for the Cheetah environment is shown in Figure \ref{fig:hero}.
Sampling and simulating the design-policy combinations in the Pareto set show that the hypernetwork is capable of learning a diverse set of skills, such as jumping to maximize height, sitting to minimize energy, and running forward to maximize speed.
As expected, designs with long legs achieve maximal height reward.
Short-legged designs run fast since lower inertia allows them to oscillate back and forth faster.
It is trivially easy for a policy to maximize the energy reward by doing nothing, so almost every design does well on this reward.

Similarly, we compute the 2-dimensional Design Pareto frontier for the Walker environment, and show it in Figure \ref{fig:paretowalker}.
Here, symmetric designs end up dominating on almost all reward combinations, except for a purely energy-maximizing strategy where one leg is much shorter than the other.
This demonstrates the capability of our algorithm to discover the relevant subspaces of an otherwise intractable design space.

\begin{figure}
    \centering
    \includegraphics[width=\linewidth]{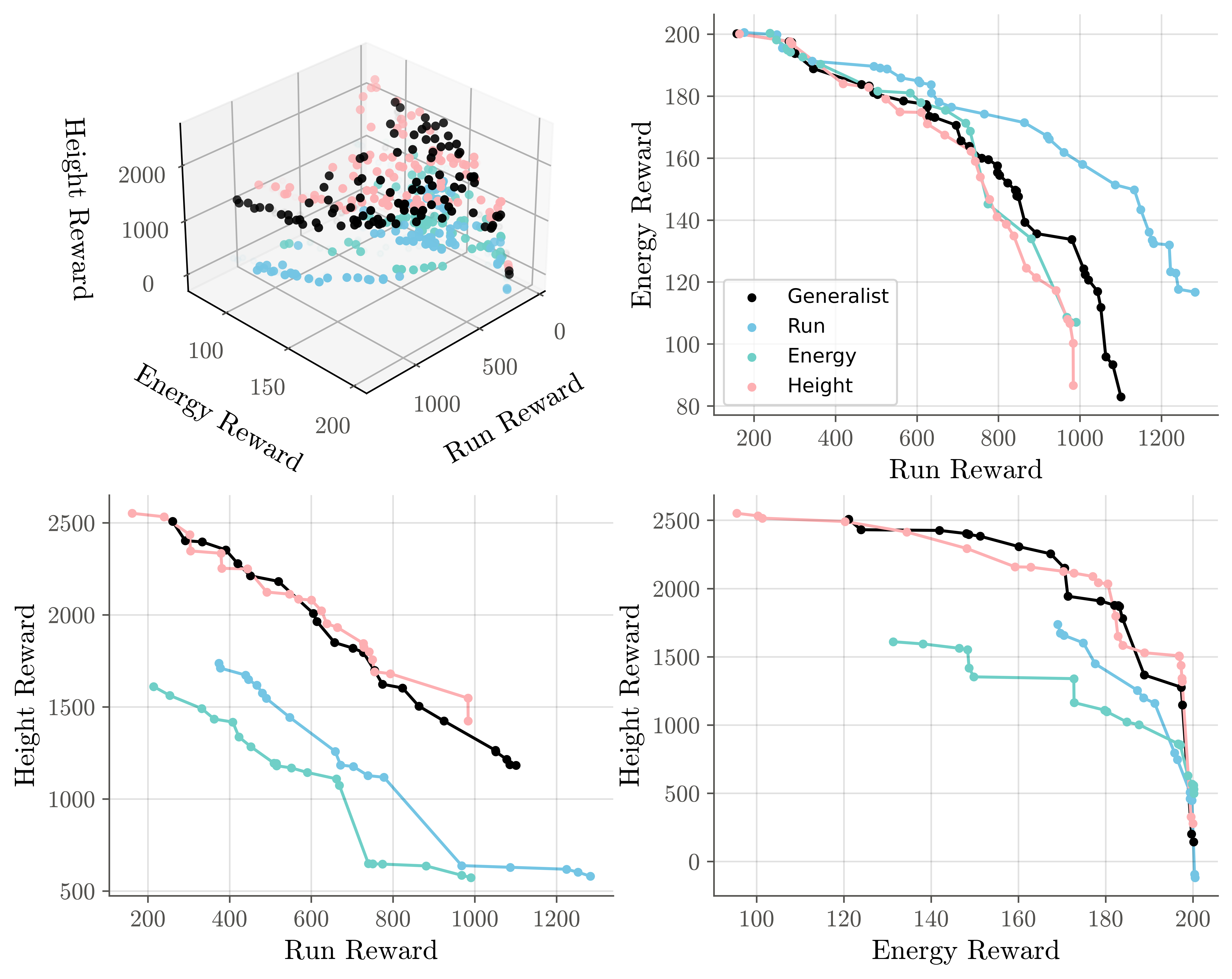}
    \caption{Fixed-Design Pareto frontiers for the generalist Cheetah, and a few selected specialists. Note that the run-optimized design far outperforms the others at running and energy savings, and the generalist design performs approximately similar to the height-optimized design.}
    \label{fig:generalist_pareto}
\end{figure}

\begin{figure*}
    \centering
    \includegraphics[width=\linewidth]{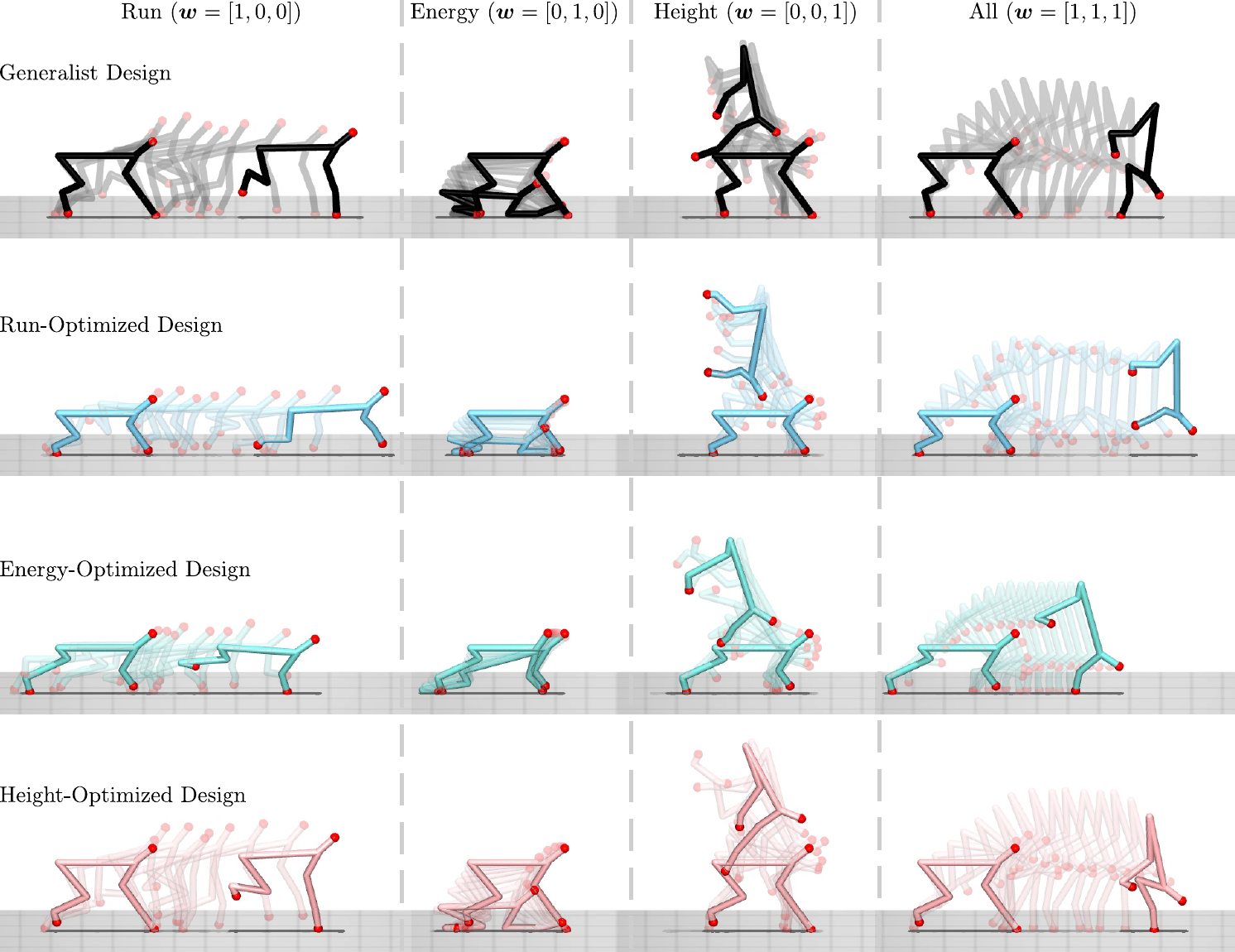}
    \caption{Procedurally-generated generalist and specialist Cheetah robots obtained using MOCHA.
    The generalist robot maximizes the hypervolume of its associated Pareto frontier, while each task-optimized design maximizes the scalar reward according to its objective prioritization.}
    \label{fig:generalist}
    \vspace{-2mm}
\end{figure*}

\subsection{Comparison to Multi-Objective Design-Conditioned MLP Policy}
In order to demonstrate the effectiveness of the hypernetwork architecture, we also train an MLP-based multi-objective universal policy (referred to as MDMLP) on the Cheetah and Walker environments.
The MDMLP policy takes in the observation and context vectors, and outputs an action.
It is comprised of 3 hidden layers, selected such that the total number of parameters is equal to that of the MDH.
The MDMLP is trained with the same multi-objective PPO loss functions as the hypernetwork.
A summary of the results of the comparison is provided in Table \ref{table:results}, with a visualization of the obtained hypervolumes against the number of environment steps provided in Figure \ref{fig:training}.

We train all networks for 3 billion timesteps.
For Cheetah, the MDH achieved a 28\% greater hypervolume than the MDMLP policy, while for the walker it achieved a 6\% greater hypervolume.
Slices of the Cheetah's Design Pareto frontiers obtained by the MDH and MDMLP are shown in figure \ref{fig:paretocheetah}.
The MDMLP performs similarly to the MDH for the Height and Energy objectives but fails at Run, demonstrating the hypernetwork's superior ability to represent a diverse set of behaviors.
Additionally, for both environments, the training time for the MDH was a third of that of the MDMLP.
This difference arises because forward inference of the  hypernetwork only occurs once per rollout, and the much smaller policy network is used every timestep.
In contrast, the large MDMLP must be used for each inference at every timestep, requiring more computation and slowing down overall throughput.

% \subsection{Optimality Gap of Hypernetwork Policies}
% \label{sec:suboptimal}
% In order for the MO-UP to be useful as a measure of design quality, its performance must be close to optimal for every design in the design space.
% To measure this, we select 6 designs and tradeoffs on the Design Pareto frontier and use those parameters to train single objective, single design policies using standard MLP networks and PPO.
% We compute the average ratio of the reward obtained by each of these policies to the scalarized reward obtained by the MO-UP with the corresponding context, and term this the \textit{optimality gap}.
% \textcolor{red}{Compute optimality gap here}.

\subsection{Comparison of Generalist and Specialists}
To demonstrate how a practitioner can analyze the Design Pareto frontier, we extract and test four morphologies of the Cheetah environment, with each fixed-design Pareto frontier shown in Figure \ref{fig:generalist_pareto}.
Specifically, we select three extreme ``specialist'' designs, each maximizing one specific objective, and one ``generalist'' design chosen as the solution to the optimization problem of Section \ref{sec:generalist}. These morphologies are illustrated in Figure \ref{fig:generalist}.
By evaluating all four morphologies across a selection of objectives, one can observe the relationship between a robot's specific physical traits, strategies taken by its control policy, and its performance on the task set.
For example, we observe that short legs help the run-optimized design not only run faster, but save energy while doing so.
Additionally, we see that the generalist design is morphologically similar and performs similarly to the height-optimized design.
The discovery of this non-obvious generalist robot is made possible through the multi-objective problem formulation of MOCHA.

% \begin{figure}
% \centering
%     \begin{subfigure}[b]{0.48\linewidth}
%         \centering
%         \includegraphics[width=\linewidth]{fig/dsup_vs_usup_run_energy.pdf}
%         \caption{}
%         \label{fig:run_energy}
%     \end{subfigure}
%     \hfill
%     \begin{subfigure}[b]{0.48\linewidth}
%         \centering
%         \includegraphics[width=\linewidth]{fig/dsup_vs_usup_run_height.pdf}
%         \caption{}
%         \label{fig:run_height}
%     \end{subfigure}
    
%     \vspace{1em} % Vertical space between rows
    
%     \begin{subfigure}[b]{0.48\linewidth}
%         \centering
%         \includegraphics[width=\linewidth]{fig/dsup_vs_usup_energy_height.pdf}
%         \caption{}
%         \label{fig:energy_height}
%     \end{subfigure}
%     \hfill
%     \begin{subfigure}[b]{0.48\linewidth}
%         \centering
%         \includegraphics[width=\linewidth]{fig/Trajectories.pdf}
%         \caption{}
%         \label{fig:trajs}
%     \end{subfigure}
    
%     \caption{(\ref{fig:run_energy}-\ref{fig:energy_height}): projections of the Design Pareto frontier across the Run, Height, and Energy objectives. (\ref{fig:trajs}): Trajectories taken by simulating designs on the Run-Height Pareto frontier.}
%     \label{fig:results}
%     \vspace{-0.5cm}
% \end{figure}

\section{Limitations}
There are a few weaknesses of our approach that can be addressed in future work.
First, the universal policy may not be the most accurate way to score a design.
If the universal policy does not discover optimal strategies, a suboptimal design may be over-rated.
While this may be resolved by training a larger hypernetwork, more studies are required to better understand the scaling capability of hypernetworks.

Second, the design space representation must be isomorphic to an $n_d$-dimensional hypercube.
This means our algorithm cannot explicitly represent complex design constraints (e.g. self-collision avoidance).
One solution to address this would be to add a cost for design constraint violations to prevent these designs from being assigned to the Pareto set.
% However, this solution still considers constraint-violating designs during training, which is not as sample-efficient as other possible solutions.

Lastly, our work does not consider sim-to-real transfer of the learned policies. 
It may be possible that some parts of the design space are more likely to exhibit good sim-to-real transfer.
Including domain randomization and a ``trust region'' in the design space are exciting directions of future research.

\section{Conclusion and Future Work}
In this paper, we introduce MOCHA, a framework for training and evaluating a Multi-Objective Design Hypernetwork, which is used in combination with multi-objective optimization as a foundation for multi-objective co-design.
This architectural choice is motivated by the recent success of hypernetworks in Multi-Objective RL and Meta RL.
Namely, we leverage the ability for hypernetworks to learn a diverse set of strategies while sharing information through a single policy representation, resulting in improved sample-efficiency.
MOCHA efficiently learns a Pareto set of policy and value network parameters for any design in a continuous design space, which provides a low-dimensional space to search when computing the optimal specialized design and policy for a given task.
Combining these specialized Pareto sets results in a Design Pareto frontier, allowing a designer to analyze how each design parameter affects task performance across the set of objectives.
Our framework also enables algorithmic computation of a generalist design, which achieves maximal performance across all objectives.
We demonstrate MOCHA across two environments, both comprised of a 6-dimensional design space and multiple objectives.
In both, the hypernetwork-based policy achieves a higher hypervolume in fewer training iterations and less wall-clock time.
Overall, this work provides a promising first step at leveraging Multi-Objective Reinforcement Learning to co-design optimal robot designs and control policies.

% \section{Acknowledgements}
% The authors used models from Anthropic and OpenAI to assist in the code implementation of MOCHA.

% \newpage
\bibliographystyle{ieeetr}    % Choose style: plain, apalike, unsrt, etc.
\bibliography{ref}     % File name without the .bib extension
\end{document}